# ARTICLE INFORMATION

**A dataset of over one thousand computed tomography scans of battery cells**

**Authors**

Amariah Condon[1], Bailey Buscarino[1], Eric Moch[1], William J. Sehnert[1], Owen Miles[1], Patrick K. Herring[1*], Peter M. Attia[1*]

**Affiliations**

1. Glimpse, 444 Somerville Avenue, Somerville, MA 02143

**Corresponding authors' email address**

Patrick K. Herring (patrick@glimp.se), Peter M. Attia (peter@glimp.se)

**Abstract**

Battery technology is increasingly important for global electrification efforts. However, batteries are highly sensitive to small manufacturing variations that can induce reliability or safety issues. An important technology for battery quality control is computed tomography (CT) scanning, which is widely used for non-destructive 3D inspection across a variety of clinical and industrial applications. Historically, however, the utility of CT scanning for high-volume manufacturing has been limited by its low throughput as well as the difficulty of handling its large file sizes. In this work, we present a dataset of over one thousand CT scans of as-produced commercially available batteries. The dataset spans various chemistries (lithium-ion and sodium-ion) as well as various battery form factors (cylindrical, pouch, and prismatic). We evaluate seven different battery types in total. The manufacturing variability and the presence of battery defects can be observed via this dataset. This dataset may be of interest to scientists and engineers working on battery technology, computer vision, or both.

# SPECIFICATIONS TABLE

| | |
|---|---|
| **Subject** | Manufacturing Engineering |
| **Specific subject area** | CT scans of lithium-ion and sodium-ion batteries for inspection of manufacturing quality |
| **Type of data** | Processed images (PNG format) |
| **Data collection** | We procured 1,015 commercially available batteries. The scans were collected using a system designed for industrial X-ray computed tomography (Nikon XT H 225 ST 2x). The system was equipped with a rotating target tungsten anode source. Data was collected using Inspect-X and reconstructed using CTPro (both version XT 6.12). The scans were then processed using Glimpse's scan post-processing software (version 0.1.0). |
| **Data source location** | All scans were acquired and processed at Glimpse's facility in Somerville, MA, USA. |
| **Data accessibility** | Repository name: Figshare<br>Data identification number: 10.25452/figshare.plus.25330501<br>Direct URL to data: https://doi.org/10.25452/figshare.plus.25330501<br>Instructions for accessing these data: Note that the full dataset is several hundred gigabytes. The data is licensed via CC-BY-NC-SA 4.0. |

# VALUE OF THE DATA

- To the best of our knowledge, this dataset is the largest publicly-available dataset of both battery manufacturing quality and industrial CT scans.
- The dataset spans seven different types of batteries, including different chemistries (lithium-ion and sodium-ion) and form factors (cylindrical, pouch, and prismatic).
- Manufacturing variability within large batches of the same cell model can be observed for two of the seven battery types. These large batches consist of 400 and 500 cells each.
- The data are provided in PNG format, which is both readily-consumable and lossless.
- The data can be used to study the manufacturing variability and quality of lithium-ion and sodium-ion batteries and to develop new computer vision routines for battery quality inspection.

# BACKGROUND

Many prior works have used CT scanning to study batteries.[1], [2], [3], [4], [5], [6], [7], [8], [9], [10], [11], [12], [13], [14], [15], [16], [17], [18] However, all of these works investigate 1-10 batteries in total. Three primary factors limit the ability of CT scanning to quickly evaluate a large number of objects. First, using suboptimal CT hardware necessitates slow scan times (typically hours). Second, CT scans generate large files (typically tens of gigabytes) that are inherently difficult to work with. In fact, CT scanning providers often mail hard drives to their customers to circumvent bandwidth limitations during upload. Finally, existing software-based CT analysis tools are designed for thorough inspection of a single scan but not rapid analysis of many scans. Using these existing tools, an operator may spend tens of minutes to load and analyze a single scan. These combined factors have made CT scanning an unrealistic option for inspection at the scale of high-throughput manufacturing.

The primary motivation of this dataset[19] is to showcase both rapid acquisition and analysis of battery cell CT scans. The dataset presented in this work contains over one thousand CT scans of seven different battery types. These batteries span lithium-ion and sodium-ion chemistries and cylindrical, pouch, and prismatic form factors. The acquisition time for most of these scans was approximately two minutes, which was enabled by a combination of optimized CT hardware and Glimpse's image post-processing techniques. Rapid analysis of this data was facilitated by both Glimpse's scan post-processing pipeline and the Glimpse Portal™ (https://app.glimp.se), a web-based interface for sharing, reviewing, and analyzing battery CT scans.

## DATA DESCRIPTION

Table I presents a summary of our dataset.

**Table I.** Summary of the dataset. For cylindrical form factors, the number in parentheses refers to the diameter and height in mm (e.g., 2170 = 21mm diameter, 70mm height; "18650" has an extraneous trailing zero).

| Producer | Battery type (cell model) | Chemistry | Form factor | Number of cells | Voxel size (µm) |
|---|---|---|---|---|---|
| EVE | INR18650/33V | Lithium-ion | Cylindrical (18650) | 400 | 14.4 |
| HAKADI | SIB18650/3V | Lithium-ion | Cylindrical (18650) | 49 | 14.4 |
| Samsung | 50E | Lithium-ion | Cylindrical (2170) | 500 | 16.4 |
| Vapcell | F56 | Sodium-ion | Cylindrical (2170) | 25 | 16.4 |
| BYD | FC4680 | Lithium-ion | Cylindrical (4680) | 25 | 35.0 |
| Tenergy | 6050100 | Lithium-ion | Pouch (51mm W x 6.0mm D x 102.5mm H) | 10 | 18.5 |
| PowerSonic | PSL-FP-IFP2770180EC | Lithium-ion | Prismatic (70mm W x 27mm D x 165mm H) | 5 | 34.0 |

The directory structure of the dataset is as follows:

1. The highest-level directory is the cell type.
2. The second-highest-level directory is the scan.
3. The third-highest-level directory is the slice orientation. We define slice orientation in detail in the next section.

4. Finally, this directory contains individual slice images corresponding to position within the cell.

All slice images are presented in PNG format, which is a lossless image codec.

# EXPERIMENTAL DESIGN, MATERIALS AND METHODS

**Sample acquisition**

1,015 commercially available battery cells were acquired from various sources, as described in Table II.

**Table II.** Sourcing information for the seven battery types.

| Producer | Battery type (cell model) | Source |
| --- | --- | --- |
| EVE | INR18650/33V | IMRbatteries.com |
| HAKADI | SIB18650/3V | selianenergy.com |
| Samsung | 50E | IMRbatteries.com |
| Vapcell | F56 | 18650batterystore.com |
| BYD | FC4680 | selianenergy.com |
| Tenergy | 6050100 | power.tenergy.com |
| PowerSonic | PSL-FP-IFP2770180EC | digikey.com |

The samples were selected for their representativeness and their diversity. The majority of the cells in this dataset are cylindrical lithium-ion cells, which are commonly used in battery-powered devices. In total, this dataset contains three different cylindrical diameter-height combinations (18650, 2170, and 4680) as well as pouch and prismatic form factors. Battery chemistry diversity is captured with the two predominant lithium-ion cathode chemistries (nickel-manganese-cobalt (NMC) and lithium-iron-phosphate (LFP)) and two different battery chemistries (lithium-ion and sodium-ion).

Some of the cell types had barcodes containing cell serial numbers, while others did not. Whenever possible, we read in the cell barcode with a barcode reader. Cells without barcodes were assigned unique serial numbers representing the order in which they were scanned. Each cell's barcode or serial number is associated with a given scan in the published dataset.

The box information is described in Table III and Table IV or the cells that were purchased in large quantities (EVE INR18650/33V and Samsung 50E). This information may be relevant for studying cell-to-cell variation.

**Table III.** Box information for the EVE INR18650/33V cells. The cells were shipped in four boxes of 100 cells each. The cells from each box were scanned before moving on to the next box.

| Box number | First cell serial number | Final cell serial number |
|---|---|---|
| 1 | LLAQ33T068462 | LLAQ21T072655 |
| 2 | KLAQ31T085151 | KLAQ34T016759 |
| 3 | KLAQ34T018447 | KLAQ31T085158 |
| 4 | KLAQ31T085116 | KLAQ34T029411 |

**Table IV.** Box information for the Samsung 50E cells. The cells from each box were scanned before moving on to the next box.

| Box number | First cell serial number | Final cell serial number | Notes |
|---|---|---|---|
| 1 | LH1T-1 | LH1T-130 | Box of 130 cells |
| 2 | LH1T-131 | LH1T-260 | Box of 130 cells |
| 3 | LH1T-261 | LH1T-390 | Box of 130 cells |
| - | LH1T-391 | LH1T-400 | Small plastic holders |
| 4 | LH1T-401 | LH1T-450 | Box of 50 cells |
| 5 | LH1T-451 | LH1T-500 | Box of 50 cells |

Upon cell receipt, the open-circuit voltage and dimensions were measured for a small subset of cells. First, the open-circuit voltage was measured via a voltmeter (Fluke 107) to confirm that the cells had not shorted during shipment. Second, the cell dimensions were measured and recorded. We found that the measured dimensions often varied substantially from the nominal dimensions; for instance, the Vapcell F56 cells had a measured diameter of 21.5 mm (vs. 21.0 mm nominal) and a measured height of 71.0 mm (vs. 70.0 mm nominal). Cell dimensions did not vary significantly within a cell type.

**Data acquisition**

Each cell was individually scanned using the Nikon XT H 225 ST 2x system in Glimpse's facility. This system was equipped with a 225 kV rotating anode X-ray source. The system was last calibrated by a qualified technician three weeks before scan acquisition. The data acquisition was performed using Nikon's Inspect-X software (version XT 6.12). Scan acquisition settings were set to balance both scan time and image quality.

**Data processing**

First, the projections were reconstructed using Nikon's CT Pro software (version XT 6.12). The reconstructed volumes were often dozens of gigabytes in size. As an example, a volume with dimensions of 1500 x 1500 x 4000 voxels and data saved as 32-bit/4-byte single-precision floats would have a reconstructed volume size of 36 GB.

Then, the reconstructed volumes were transferred to another computing system (the "GlimpseBox") and processed through Glimpse's scan post-processing pipeline. This imaging pipeline performs several image enhancement steps, including intensity adjustment, cropping, and denoising. These processing steps were consistent across all scans of the same form factor (i.e., cylindrical, pouch, and prismatic).

One of the last steps in this pipeline is slicing, in which the 3D volume can be "sliced" in various orientations to reveal a 2D cross section in a particular orientation. These cross sections allow for viewing of the object's interior. The slice orientations are defined by their in-place dimension. For cylindrical cells, the primary axis of a radial slice is the radius, and the primary axis of an axial slice is the cylindrical axis, as depicted in Figure 1. $Z=0$ refers to the bottommost radial slice of a cell, while $\vartheta=0$ is arbitrary (depends on the orientation of the cell as it was inserted into the scanner).

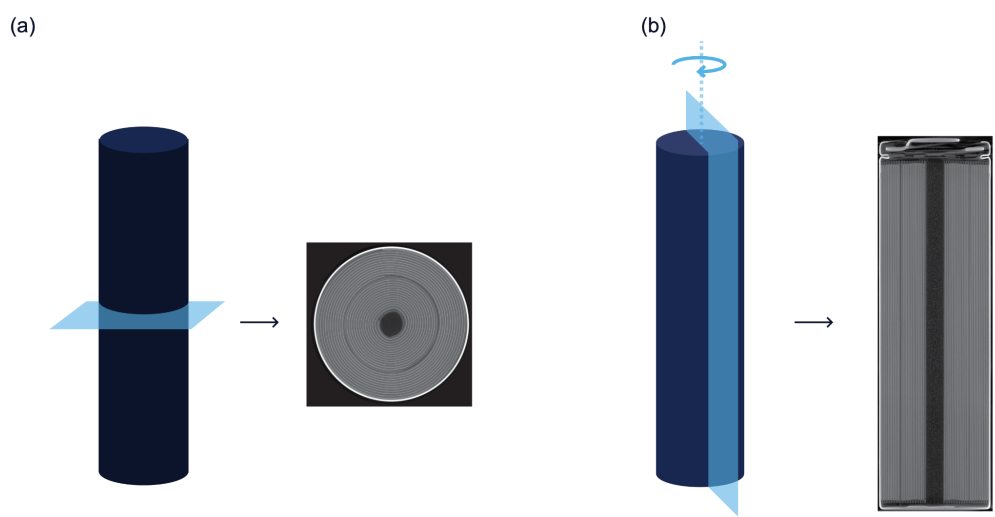

**Figure 1.** Depiction of (a) radial slices and (b) axial slices for cylindrical cells.

For pouch and prismatic cells, the three possible orientations are xy, xz, and yz (Figure 2). The two letters refer to the two in-plane dimensions. For all three orientations, $Z=0$ refers to the bottommost slice of the cell.

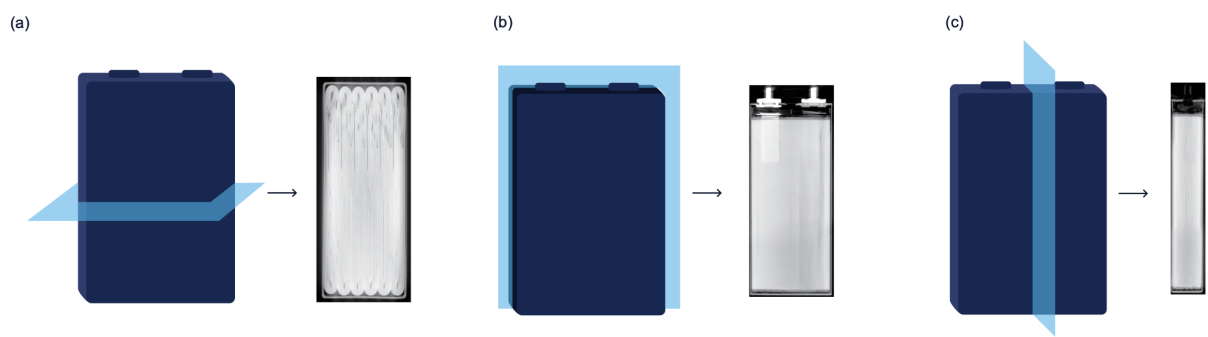

**Figure 2.** Depiction of (a) xy, (b) xz, and (c) yz slices for pouch and prismatic cells.

The slice positions were tuned such that the jellyroll (i.e., a region with low variation with respect to position) was coarsely sliced, and regions outside of the jellyroll such as the cell header (i.e., a region with low variation with respect to position) were finely sliced.

**Data review**

After reviewing the scans, we found that four of the procured sodium-ion cells were in fact lithium-ion cells, as identified via the presence of copper foil on the anode. These cells appeared visually identical to the others in the batch. These four cells had very poor quality, i.e., anode-cathode overhang violations. These counterfeit cells are present in the sodium-ion dataset.

# LIMITATIONS

The scans were generated in batches over a period of multiple days. As such, some variation in image quality and measurement accuracy may have occurred due to differences in daily procedures such as X-ray source conditioning, manipulator homing, and shading corrections.

The dataset is not a representative sampling of all lithium-ion and sodium-ion batteries produced today. As such, users should avoid over-generalizing this dataset to all battery designs (i.e., for training AI models).

All CT scans have artifacts that may interfere with scan interpretation.[20] In particular, beam hardening artifacts and metal streaking artifacts are clearly visible in some of these scans.

The raw data is not hosted as the dataset size would be several dozen terabytes.

The software post-processing pipeline is not publicly available as it is proprietary to Glimpse.

On the Glimpse Portal™, users can view the results of computer vision algorithms for automated inspection. These algorithms detect battery defects and measure key battery properties. The results of these algorithms are not included in this dataset.

# ETHICS STATEMENT

The authors have read and followed the ethical requirements for publication in Data in Brief and confirm that the current work does not involve human subjects, animal experiments, or any data collected from social media platforms.

# CRediT AUTHOR STATEMENT

Amariah Condon: Data Curation, Investigation, Project administration, Resources, Validation; Bailey Buscarino: Investigation, Software; Eric Moch: Investigation, Validation; William J. Sehnert: Investigation, Validation; Owen Miles: Investigation; Patrick K. Herring: Data Curation, Investigation, Project administration, Software, Validation; Peter M. Attia: Conceptualization, Data Curation, Investigation, Resources, Software, Supervision, Writing – Original Draft, Validation, Visualization

# ACKNOWLEDGEMENTS

This research did not receive any specific grant from funding agencies in the public, commercial, or not-for-profit sectors.

# DECLARATION OF COMPETING INTERESTS

The authors declare the following financial interests which may be considered as potential competing interests: All authors are co-founders or founding engineers with a financial interest in Glimpse Engineering Inc., which specializes in high-throughput CT scanning for batteries.